\documentclass{article}

\usepackage{geometry}
\usepackage{indentfirst}
\usepackage{graphicx}
\usepackage{amssymb}
\usepackage{booktabs}
\usepackage{caption}
\usepackage{soul}
\usepackage{xcolor}
\usepackage{subcaption}
\usepackage{multirow}
\usepackage{pdflscape}
\usepackage{float}
\usepackage{amsmath}
\usepackage{tabularx}
\usepackage{longtable}
\usepackage{standalone}
\usepackage{tikz}
\usepackage{lineno}
\usepackage{calc}

\usepackage[ruled,vlined]{algorithm2e}
\usepackage[breaklinks,colorlinks]{hyperref}

\newcommand{\figref}[2]{%
    Fig.\hyperref[#1]{~\ref{#1}#2}%
}

\newcommand{\extfigref}[2]{%
Extended Data Fig.\hyperref[#1]{~\ref{#1}#2}%
}

\renewcommand{\figurename}{Fig.}

\graphicspath{ {./images/} }

\begin{document}

    \title{CellSeg1: Robust Cell Segmentation with One Training Image}

    \author{
        Peilin Zhou\textsuperscript{1},
        Bo Du\textsuperscript{1},
        Yongchao Xu\textsuperscript{1} \\
        \vspace{1em} \\
        \textsuperscript{1} School of Computer Science, Wuhan University. \\
        \vspace{1em} \\
    }
    \date{}
    \maketitle
    \begin{abstract}
        Recent trends in cell segmentation have shifted towards universal models to handle diverse cell morphologies and imaging modalities. However, for continuously emerging cell types and imaging techniques, these models still require hundreds or thousands of annotated cells for fine-tuning. We introduce \textit{CellSeg1}, a practical solution for segmenting cells of arbitrary morphology and modality with a few dozen cell annotations in 1 image. By adopting Low-Rank Adaptation of the Segment Anything Model (SAM), we achieve robust cell segmentation. Tested on 19 diverse cell datasets, CellSeg1 trained on 1 image achieved 0.81 average mAP at 0.5 IoU, performing comparably to existing models trained on over 500 images. It also demonstrated superior generalization in cross-dataset tests on TissueNet. We found that high-quality annotation of a few dozen densely packed cells of varied sizes is key to effective segmentation. CellSeg1 provides an efficient solution for cell segmentation with minimal annotation effort.
    \end{abstract}

    \section{Introduction}

    Cell segmentation is a critical step in quantitative cell biology, essential for measuring various cellular characteristics such as morphology, position, RNA expression, spatial transcriptomics, and protein expression~\cite{caicedo2017data, chen2023scs,petukhov2022cell}. However, the vast diversity of cell types across different species and organisms, coupled with variations in imaging methods and equipment, results in a wide range of cellular morphologies and colors~\cite{NeurIPS-CellSeg, bannon2021deepcell, vicar2019cell, sekh2021physics}. This makes universal automatic cell segmentation a very challenging task~\cite{moen2019deep, shrestha2023efficient}.
    In addition, cells may be tightly packed with indistinct boundaries, further complicating the segmentation process~\cite{weigert2018content,amat2014fast}.

    Deep learning approaches, particularly U-Net-based models, have significantly advanced medical image segmentation~\cite{unet,isensee2021nnu}. In the field of cell segmentation, StarDist~\cite{stardist} emerged as a pioneering method, employing star-convex~\cite{isack2018k} representations to accurately capture cell contours for star-convex cells. To distinguish adjacent cells, researchers have proposed various vector field-based methods for overlapping regions. HoverNet~\cite{hovernet} predicts horizontal and vertical distances to cell mass centers, simultaneously performing cell detection, segmentation, and classification in pathology images. Cellpose 1.0~\cite{cellpose} utilizes a heat diffusion model to represent cell shapes, effectively handling cells of different scales and morphologies, especially neurons with elongated axons and dendrites. Omnipose~\cite{omnipose} specializes in segmenting rod-shaped bacteria. The Location-sensitive deep learning method~\cite{lsd} predicts distances from sphere centers to the centroids of sphere-cell intersection regions for neuronal segmentation.

    These methods are mostly tailored to specific cell image characteristics, often requiring careful model tuning based on the target cell type and extensive data annotation. Recent trends have shifted towards universal models~\cite{cellpose2, NeurIPS-CellSeg, ma2023towards, cellsam, medsam}. In particular, Cellpose 2.0~\cite{cellpose2} introduces a general cellpose-cyto2 model trained on diverse cell types. This approach combines a pre-trained universal model with a human-in-the-loop fine-tuning strategy. For new imaging modalities similar to those in the training set, this method requires less annotations, often needing only a few hundreds to a thousand annotated cells for effective adaptation. However, challenges persist in practical implementation. When encountering cell types or imaging modalities significantly different from the original training data, the model's performance may degrade substantially~\cite{NeurIPS-CellSeg, omnipose}, necessitating more annotation efforts. These limitations underscore the ongoing challenge of developing truly adaptable and efficient cell segmentation methods across diverse biological contexts.

    In this work, we propose a novel approach called \textit{CellSeg1} that dramatically reduces the annotation burden to just 1 training image with a few dozen cells, while maintaining robust segmentation performance across diverse cell types and imaging modalities. CellSeg1 explores the generalization ability of the Segment Anything Model (SAM)~\cite{sam} and employs Low-Rank Adaptation (LoRA)~\cite{lora} for efficient fine-tuning. Remarkably, CellSeg1 achieves an average $mAP_{0.5}$ (mean average precision at 0.5 intersection over union) of 0.81 across 19 datasets of different cell types and imaging modalities using only a single annotated image containing several dozen cells~(\figref{fig:figure_1}{a}, \figref{fig:figure_5}{a}). We conducted comprehensive experiments on a wide range of cell morphology datasets to validate our CellSeg1's robustness and generalization ability. Furthermore, we performed an in-depth analysis of the impact of training image annotation quality, providing valuable guidelines for data annotation in practical applications with CellSeg1. To facilitate widespread adoption, we have developed a user-friendly graphical interface that simplifies the process of training, testing, and visualizing results~(\extfigref{fig:extended_figure_6}{}). Through careful optimization, our implementation enables model training and inference on a single consumer-grade GPU with 8 GB VRAM~(\extfigref{fig:extended_figure_5}{}), making it accessible to most research laboratories.

    \section{Results}
    \subsection{CellSeg1 achieves outstanding performance using 1 training image}
    To evaluate CellSeg1's performance, we utilized diverse datasets encompassing various cell segmentation challenges (Supplementary Table 1). The 2018 Data Science Bowl (DSB)~\cite{dsb2018} dataset tests adaptability to scale variation, with cell areas ranging from 21 to 6475 pixels. The Cellpose Specialized (CPS)~\cite{cellpose} dataset demands precise segmentation of neurons with complex dendritic structures. The Escherichia Coli Brightfield (ECB)~\cite{deepbacs} dataset requires accurate detection of bacterial cell division states to determine whether dividing cells should be segmented as one or two objects. The Bacillus Subtilis Fluorescence (BSF)~\cite{deepbacs} dataset contains densely packed bacterial chains and microcolonies with extensive cell-cell contacts. The CellSeg-Blood (CSB)~\cite{NeurIPS-CellSeg} dataset uses RGB color images rather than separate nuclei and cytoplasmic channels. The Cellpose generalized (CPG)~\cite{cellpose} dataset features highly variable cell appearances, challenging the model's ability to handle diverse morphological characteristics. The TissueNet (TSN)~\cite{tissuenet} dataset, comprising 14 tissue subsets from various imaging platforms, contains inherent domain gaps despite visual similarities to human observers, testing deep learning models' generalization capabilities.

    We evaluated CellSeg1 against several state-of-the-art cell segmentation approaches, each employing a different training paradigm. StarDist~\cite{stardist} trains on the complete dataset of similar images, while Cellpose-cyto2~\cite{cellpose2} employs a two-stage strategy: first pre-training on diverse cell images (including CPG training dataset) to create a universal model, followed by fine-tuning on the complete dataset of similar images. CellSAM~\cite{cellsam} attempts to replicate SAM's success by following its large-scale training approach, utilizing over 1 million cell annotations (from TissueNet~\cite{tissuenet}, Cellpose~\cite{cellpose}, DSB~\cite{dsb2018}, DeepBacs~\cite{deepbacs}, and many other datasets) for training. Note that we directly report the performance of pre-trained CellSAM in all benchmarks. We adopted $mAP_{0.5}$ as the primary evaluation metric, consistent with the Cellpose benchmark~\cite{cellpose}.

\begin{figure}[H]
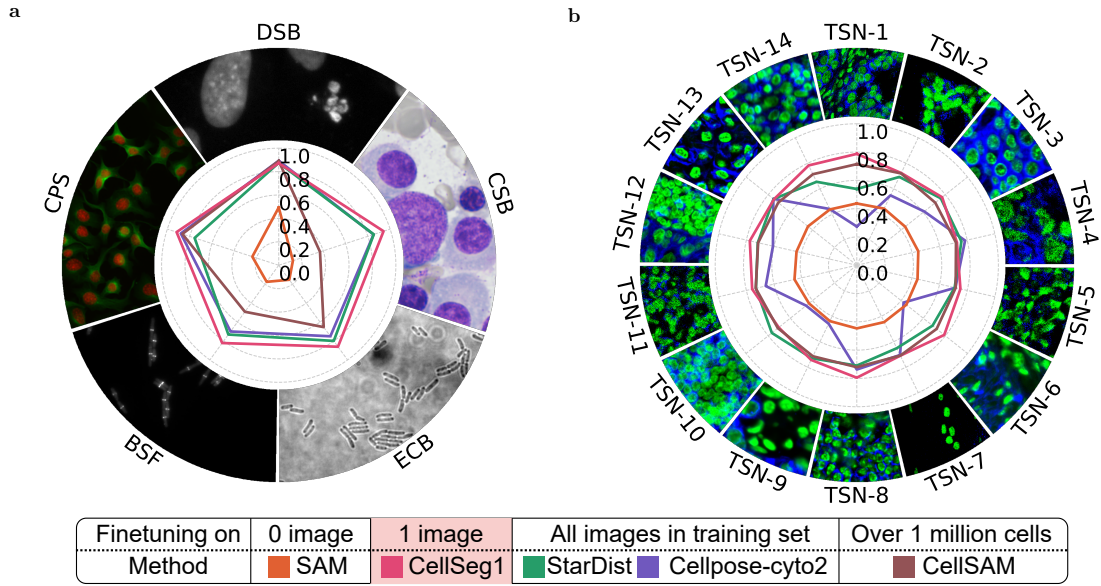

        \includestandalone[width=\textwidth]{figures/figure_1/figure_1}
        \caption{CellSeg1 trained on just a single image outperforms existing methods trained on extensive images. (a) Performance comparison on 5 diverse datasets, where inner radar charts show $mAP_{0.5}$ scores and outer rings show the single training image used for CellSeg1 to achieve the corresponding inner radar chart results. (b) Cross-subset generalization test on 14 TissueNet nuclei subsets, where generalization ability is assessed using leave-one-out cross-validation across subsets.}
        \label{fig:figure_1}
    \end{figure}

    We conducted our experimental evaluation in two complementary ways~(\figref{fig:figure_1}{a,b}). In the first evaluation~(\figref{fig:figure_1}{a}), we trained CellSeg1 on each dataset using its best-performing single training image. The detailed performance analysis using different single training images is presented separately in~\figref{fig:figure_3}{}. For the second evaluation, we assessed cross-domain generalization across the TissueNet nuclei dataset using a more practical approach: for each subset, we selected the first image (based on alphabetical filename sorting) as the training image for CellSeg1, while benchmark methods were trained on their complete training sets~(\figref{fig:figure_1}{b}).

    CellSeg1, trained on just a single image, demonstrated superior performance compared to other methods trained on extensive datasets~(including the image used to train CellSeg1) in four out of five datasets, with DSB being the only exception~(\figref{fig:figure_1}{a}). While CellSAM outperformed StarDist on CPS and DSB, it consistently underperformed compared to Cellpose-cyto2 across all datasets. Notably, CellSAM's performance was constrained by its training strategy, which relied on separate nuclei and cellular channels~(Supplementary Table 2), leading to poor results when testing on grayscale-converted CSB images. In contrast, while Cellpose-cyto2 also using separate nuclei and cellular channels, its fine-tuning on grayscale CSB data yielded better results. Both CellSeg1 and StarDist demonstrated greater versatility by working directly with original RGB images.

    Our CellSeg1's effectiveness is further validated by its cross-dataset performance, where single-image trained models outperformed methods trained on complete datasets in 11 out of 14 TissueNet subsets~(\figref{fig:figure_1}{b}). This is particularly significant because in real-world applications, subtle variations in imaging equipment and acquisition protocols can create domain gaps that are imperceptible to human observers but challenging for deep learning models~\cite{guan2021domain, stacke2020measuring}. CellSeg1's robust cross-domain generalization ability eliminates the need for re-annotation when working with similar imaging conditions, substantially reducing annotation effort.

    \subsection{CellSeg1 leverages pretrained features for efficient cell segmentation}

    Recent approaches applying SAM~\cite{sam} to cell segmentation have shown promise, but often require extensive additional trainable parameters~\cite{zhang2023input,cellsam,cellvit, samed, cai2024biosam, chen2024sam, samformicro}. Methods such as training separate prompt generation networks~\cite{cellsam, cai2024biosam} or retraining the decoder~\cite{cellvit, samed,chen2024sam} typically introduce a large number of trainable parameters, necessitating substantial annotated data for overfitting prevention~\cite{nakkiran2021deep, lora, kaplan2020scaling, houlsby2019parameter}.
    However, the morphological diversity of cellular images renders the annotation of large datasets for each cell type both time-consuming and cost-prohibitive.

    Recognizing that SAM has been pretrained on a vast corpus of natural images, thereby encapsulating rich feature representations, we developed CellSeg1 to fully utilize these pretrained features while minimizing the demand for annotated samples. Our CellSeg1 employs LoRA~\cite{lora}, introducing only a small number of trainable parameters in SAM's encoder and decoder. This strategy not only preserves SAM's pretrained weights but also significantly reduces the number of trainable parameters, effectively mitigating the risk of overfitting under limited-sample conditions.

\begin{figure}[H]
        \centering
        \includegraphics[width=1.0\textwidth]{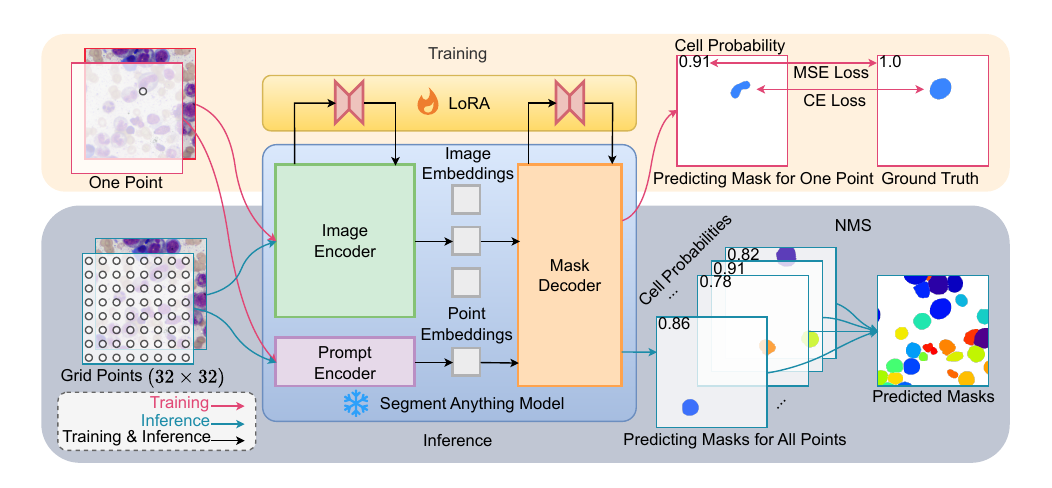}
        \caption{The pipeline of CellSeg1 for cell segmentation. The pipeline consists of two main stages: Training (top) and Inference (bottom). During training, the model uses LoRA to fine-tune the SAM on a single point. The inference stage utilizes a grid of points to generate multiple masks, which are then filtered using Non-Maximum Suppression (NMS).}
        \label{fig:figure_2}
    \end{figure}

    The workflow of CellSeg1 comprises two primary phases: training and inference~(\figref{fig:figure_2}{}). During training, the model receives paired image and point inputs. For positive sample points located within cells, the model learns to predict the corresponding cell mask and output a cell probability of 1.0. For negative sample points in background regions, the model is trained to output a cell probability of 0.0. We reinterpret SAM's original output for predicting Intersection over Union (IoU) as cell probability, indicating whether the current point is located within a cell. This learning strategy simultaneously performs segmentation and cell/background classification.

    In the inference phase, CellSeg1 leverages SAM's automatic prompt generation mode (``everythin'' mode), uniformly sampling $32 \times 32$ grid points on the input image. For each point, the model generates a mask and an associated cell probability. Through an optimized Non-Maximum Suppression (NMS) algorithm, we integrate all predicted masks to generate the final instance segmentation result.

    \subsection{CellSeg1 is robust to variations in the selected single training image}
    To rigorously assess the performance stability of CellSeg1 under single-image training conditions, we conducted a series of experiments across five representative datasets: CSB~\cite{NeurIPS-CellSeg}, ECB~\cite{deepbacs}, BSF~\cite{deepbacs}, CPS~\cite{cellpose}, and DSB~\cite{dsb2018}. The experimental design was as follows: For the CSB, ECB, BSF, and CPS datasets, we exhaustively utilized every image in each training set. Given the larger scale of the DSB dataset, we randomly sampled 100 training images. In each experiment, the model was trained using only a single image and subsequently evaluated on the complete test set. For benchmarks, we also incorporated the randomly initialized cellpose-scratch model for comparison.

\begin{figure}[H]
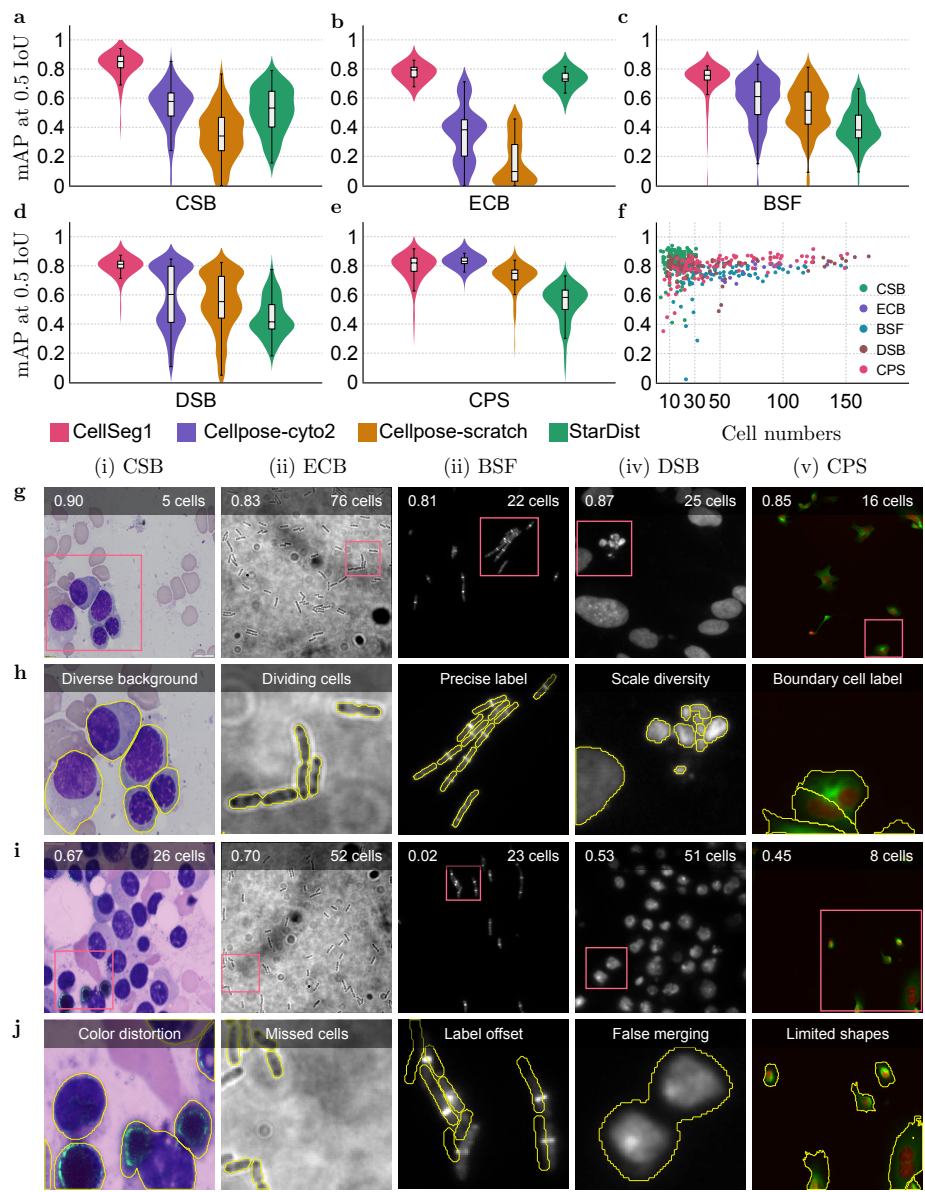

        \centering
        \includestandalone[width=0.83\textwidth]{figures/figure_3/figure_3}
        \caption{Performance analysis of CellSeg1 trained on different single training images.
            (a-e) Distribution of $mAP_{0.5}$ for each method when trained on different single images from the respective datasets. Boxes centered on medians, whiskers from Q1 - 1.5 IQR to Q3 + 1.5 IQR, constrained to the data range. IQR: interquartile range (Q3-Q1).
            (f) Scatter plot showing the relationship between the number of cells in each training image and the resulting $mAP_{0.5}$ for CellSeg1. Each point represents a single experiment using one training image.
            (g) Examples of high-quality training images that led to good performance for CellSeg1.
            (h) Zoomed-in views of the red boxed areas in (g), highlighting key features of effective training images.
            (i) Examples of low-quality training images that resulted in poor performance for CellSeg1.
            (j) Zoomed-in views of the red boxed areas in (i), illustrating characteristics of problematic training images.}
        \label{fig:figure_3}
    \end{figure}

    CellSeg1 demonstrated superior performance across five datasets compared to benchmark methods, achieving median $mAP_{0.5}$ scores above 0.8 in three datasets~(\figref{fig:figure_3}{a,d,e}). It showed better accuracy and stability in four datasets, as evidenced by concentrated violin plot distributions. On the CPS dataset~(\figref{fig:figure_3}{e}), while CellSeg1 outperformed Cellpose-scratch, it was slightly worse than Cellpose-cyto2, which was expected since CPS was included in Cellpose-cyto2's pre-training dataset. Overall, CellSeg1's consistent performance demonstrates its robust and generalizable cell segmentation capabilities.

    Analysis of the relationship between annotation quantity and model performance revealed that for several datasets including CSB, DSB, CPS, and BSF, annotating with 1 image less than 30 cells is sufficient to achieve an $mAP_{0.5}$ of 0.8~(\figref{fig:figure_3}{g}, \extfigref{fig:extended_figure_1}{a,c-e}). The ECB dataset required more extensive annotation (76 cells) to reach 0.83 $mAP_{0.5}$, primarily due to the challenge of determining whether dividing cells should be annotated as one or two instances.

    High-quality training images demonstrate several key characteristics~(\figref{fig:figure_3}{g,h}, \extfigref{fig:extended_figure_2}{}). The ideal training images contain densely distributed cells with natural overlapping patterns, offering rich contextual information. The annotations follow rigorous standards, particularly crucial for dividing cells and complex cellular structures. Clear definition of cell boundaries within clusters and representation of cells at multiple scales strengthen the training dataset. Moreover, careful annotation of cells at image edges ensures complete spatial coverage.

    In contrast, low-quality training images exhibit various deficiencies~(\figref{fig:figure_3}{i,j}). These include color distortion, incomplete or misaligned annotations, incorrectly merged adjacent cells, and insufficient morphological diversity in the selected samples. Such quality issues can substantially reduce the model's ability to learn robust features for accurate cell segmentation.

    These results suggest that \textbf{annotation quality is more important than annotation quantity} for CellSeg1. A single well-annotated image containing diverse, precisely labeled cells demonstrated superior performance compared to images containing more cells with suboptimal annotations. This finding has important implications for addressing the challenge of limited training data in practical biomedical image analysis, suggesting that efforts should focus on obtaining high-quality annotations rather than maximizing the number of training samples.

    \subsection{A single training image is enough for efficient cell segmentation}

\begin{figure}[H]
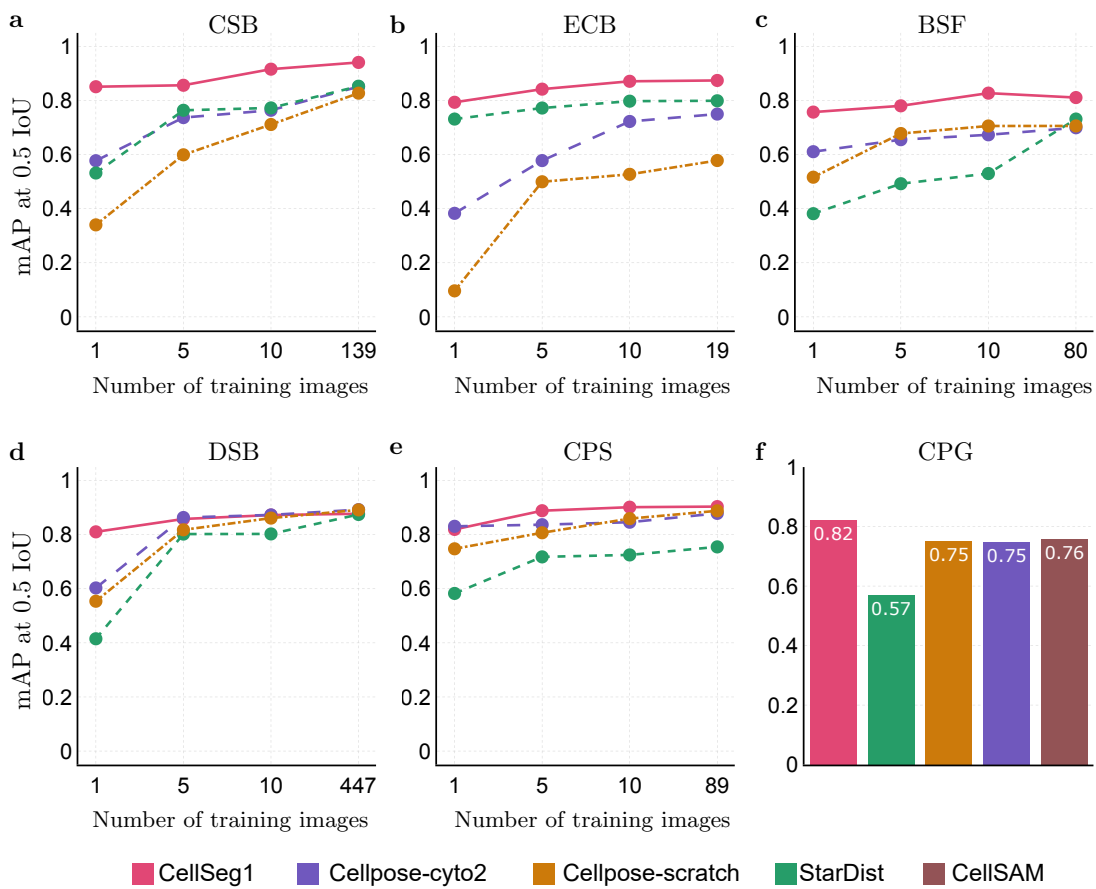

        \centering
        \includestandalone[width=\textwidth]{figures/figure_4/figure_4}
        \caption{Evaluation of model performance with varying numbers of training images. (a-e) $mAP_{0.5}$ for DSB, CSB, ECB, BSF, and CPS datasets using 1, 5, 10, and all available training images. For 5 and 10 image sets, images were selected by sorting filenames and choosing the first 5 or 10. Single-image results represent the median values from the single-image experiments shown in Figure 3. (f) Performance comparison on the CPG dataset. While other methods were trained using the complete CPG training set, CellSAM results were obtained directly using its pre-trained model (trained on over 1 million cells including Cellpose data) without additional fine-tuning.}
        \label{fig:figure_4}
    \end{figure}

    To further explore the impact of training sample quantity on CellSeg1's performance, we trained models using 5, 10, and all available training images~(\figref{fig:figure_4}{a-e}). Our experimental results revealed a striking phenomenon: while increasing the number of annotated images led to improved segmentation performance across all methods, CellSeg1 showed marginal gains compared to the substantial improvements observed in comparative methods. In fact, CellSeg1 trained on a single image already performs well, being comparable with or even better than other methods trained on extensive number of images. This finding underscores CellSeg1's advantage in data efficiency, where a single annotated training image is enough to achieve satisfied result.

    Further analysis showed that Cellpose methods exhibited excellent performance on certain datasets. For instance, on the DSB dataset, only 5 training images were needed to achieve an $mAP_{0.5}$ of 0.86. On the CPS dataset, even with random initialization, single-image training achieved an $mAP_{0.5}$ of 0.75. However, when faced with cell types significantly different from the pre-training data (such as CSB, ECB, BSF datasets), cellpose-cyto2's performance dropped substantially in few-shot scenarios. Even with increased training data, cellpose-cyto2 still underperformed compared to CellSeg1. In contrast, CellSeg1, leveraging SAM's universal visual feature extractor, demonstrated superior adaptability, effectively handling diverse cell morphologies and colors across multiple datasets.

    To further validate CellSeg1's adaptability to diverse cell types, we trained on the CPG dataset, which encompasses multiple image types~(\figref{fig:figure_4}{f}). The results showed that CellSeg1, using all training images, achieved an $mAP_{0.5}$ of 0.82, significantly outperforming comparison methods. This outcome further confirms CellSeg1's superiority in handling cells of varying morphologies.

    While specialized methods may achieve excellent results in specific scenarios through extensive data training, CellSeg1's ability to perform robustly across diverse datasets with just a single training image demonstrates the potential of leveraging large-scale pre-trained models for specialized biomedical tasks. This approach may pave the way for more generalized and efficient solutions in cell image analysis, reducing the need for extensive dataset-specific annotations and accelerating scientific discovery in cell biology.

    \subsection{CellSeg1 demonstrates strong generalization ability}
    The diverse morphologies of cells and varying imaging conditions present significant challenges to the generalization capabilities of deep learning models in cell segmentation tasks. To evaluate CellSeg1's performance in this aspect, we conducted a systematic study on the TissueNet nuclei dataset. This dataset comprises multiple subsets, each derived from different tissues and imaging modalities, providing an appropriate setting for assessing cross-domain generalization capabilities.

\begin{figure}[H]
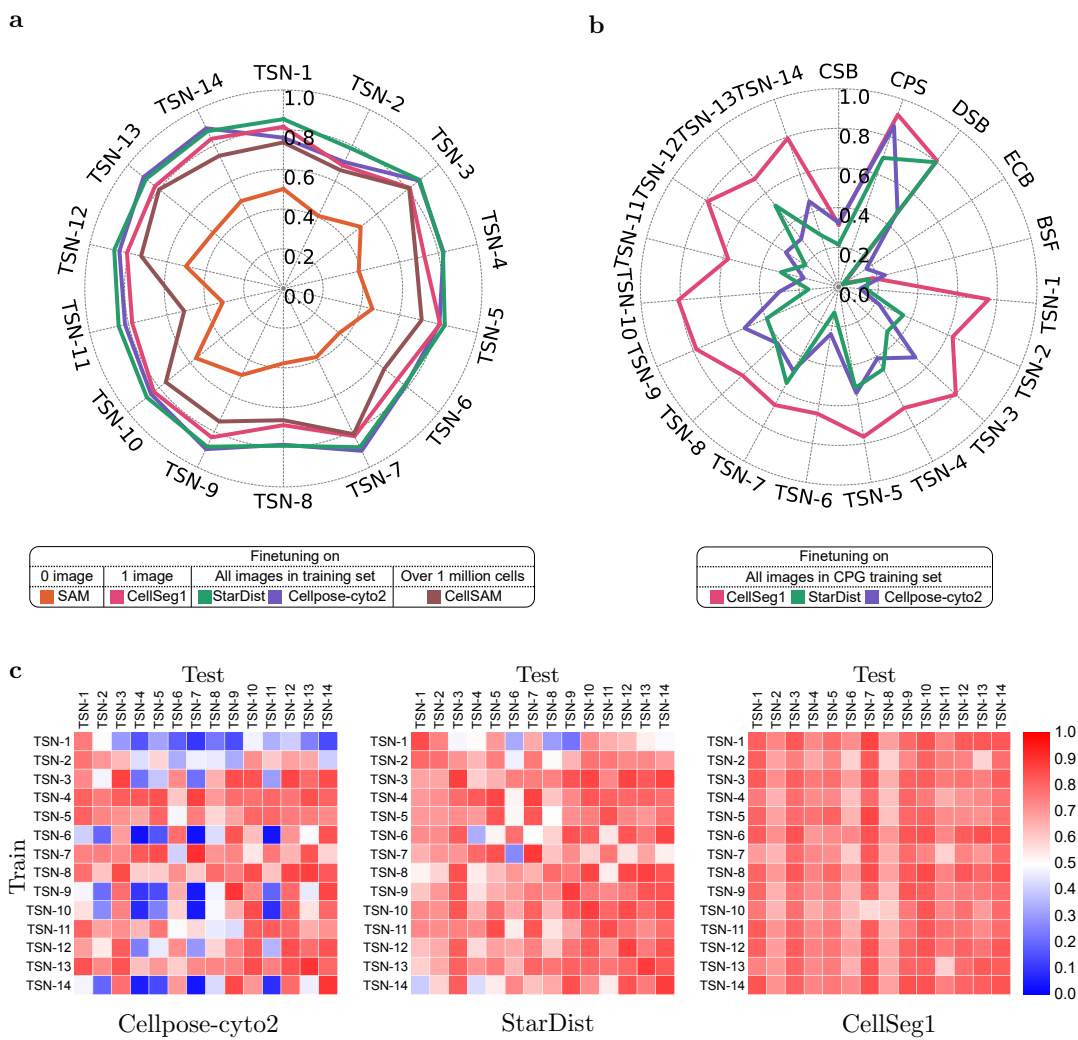

        \centering
        \includestandalone[width=\textwidth]{figures/figure_5/figure_5}
        \caption{Evaluation of CellSeg1's performance and generalization capability across diverse cell types.
            a) Comparison of model performance on 14 TissueNet subsets. CellSeg1, trained on a single image, achieves comparable performance to Cellpose-cyto2 and StarDist trained on all images, while outperforming zero-shot SAM or CellSAM.
            b) Cross-dataset generalization evaluation. Models trained on the CPG dataset were tested on various cell types. All images were converted to grayscale for Cellpose-cyto2.
            c) Detailed cross-subset generalization analysis within TissueNet. Heat maps show the $mAP_{0.5}$ scores when training on one subset (rows) and testing on another (columns). CellSeg1, trained on a single image per subset, exhibits stronger generalization compared to Cellpose-cyto2 and StarDist trained on all images, as evidenced by the more uniformly high scores across the matrix.}
        \label{fig:figure_5}
    \end{figure}

    We first assessed CellSeg1's performance under single-image training conditions~(\figref{fig:figure_5}{a}). The results revealed that while the original SAM model achieved only approximately 0.4 $mAP_{0.5}$ across various subsets, CellSeg1 attained around 0.8 $mAP_{0.5}$ using just a single training image, approaching the performance of benchmark methods trained on complete datasets containing up to 587 images.

    To delve deeper into the model's generalization capabilities, we created a heatmap of cross-subset generalization performance for each TissueNet subset~(\figref{fig:figure_5}{c}). In this experiment, we trained the model on each subset and tested it on the others. The heatmap illustrates CellSeg1's superior cross-subset generalization ability. Even under the condition of single-image training, CellSeg1 outperformed StarDist and Cellpose-cyto2 trained on full datasets in most cross-subset scenarios. This generalization capacity suggests that CellSeg1 can adapt to subtle variations in imaging conditions and cellular morphologies without re-annotation and re-training, significantly enhancing its practical value.

    To extend our analysis of generalization, we also trained models on the CPG~\cite{cellpose} dataset, a diverse collection containing over 70,000 annotated cells across various imaging conditions, specifically curated by Cellpose for developing generalist cell segmentation models. While CellSeg1 outperformed benchmark methods on most datasets~(\figref{fig:figure_5}{b}), all methods showed notably reduced performance on CSB (RGB images), ECB and BSF (bacterial cells), datasets that differ substantially from CPG in imaging modality or cell morphology. Notably, when tested on TissueNet, CellSeg1 achieved better results using single-image training on each subset~(\figref{fig:figure_5}{a}) compared to CellSeg1 trained on the entire CPG dataset~(\figref{fig:figure_5}{b}).
    This finding reveals a potential limitation of universal cell segmentation models. Training on carefully curated diverse datasets like CPG may not perform as well as training on just one well-annotated image. This demonstrates the advantage of our CellSeg1's paradigm for rapid adaptation to any cell type in practical usage.

    \section{Discussion}
    Our study demonstrates that CellSeg1, leveraging the SAM and LoRA, achieves superior cell segmentation performance with only one training image across diverse cell types and imaging conditions. This approach significantly reduces annotation costs while maintaining strong generalization capabilities, addressing a critical challenge in practical biomedical image analysis.

    Our findings question both the necessity and practicality of pursuing the development of a highly challenging universal cell segmentation model.
    Despite extensive pre-training on diverse datasets, methods like Cellpose-cyto2 and CellSAM often struggle with novel cell types or imaging conditions, due to the inherent heterogeneity in cell morphologies and imaging modalities. Rather than pursuing a one-size-fits-all solution, CellSeg1's approach of rapid adaptation using a single high-quality annotated image presents a more practical strategy, effectively balancing model generalization and specialization through SAM's robust features and LoRA's efficient fine-tuning.

    We found that annotation quality, rather than quantity, is the primary determinant of CellSeg1's performance. This phenomenon can be attributed to two factors. Firstly, the use of Parameter-Efficient Fine-Tuning (PEFT)~\cite{houlsby2019parameter} methods like LoRA~\cite{lora} introduces fewer trainable parameters, reducing the risk of overfitting even with limited data. Secondly, in small-sample scenarios, even a few annotation errors can significantly impact performance due to their proportionally large influence. Our results indicate that as few as 30 well-annotated cells can achieve an $mAP_{0.5}$ of 0.8, with diminishing returns beyond this threshold.

    Previous research has emphasized the challenge of overlapping cells in segmentation tasks, often proposing powerful vector field-based solutions~\cite{stardist, cellpose, hovernet, lsd, omnipose, cellvit} or suggesting the use of bounding box prompts~\cite{cellsam} with SAM to avoid ambiguous cell boundaries. Surprisingly, our study reveals that CellSeg1, using only point prompts, effectively addresses the issue of overlapping cells. This capability may stem from SAM's training process, which includes the generation of ambiguous masks~\cite{sam}, potentially endowing it with an inherent ability to differentiate overlapping structures.

    Recent advances in image segmentation, particularly the memory bank approach~\cite{cheng2024putting} demonstrated in SAM2~\cite{ravi2024sam2} for video tracking, suggest a promising direction for extending CellSeg1 to three-dimensional cell segmentation~\cite{shen2024interactive, guo2024sam2point}. Just as SAM2 leverages the temporal continuity between video frames, this extending direction could effectively utilize the spatial similarity between consecutive slices in 3D cell imaging, where adjacent sections share highly similar cellular features and structures.

    \section{Methods}
    \subsection{Datasets}

    \textbf{Cellpose}. The Cellpose dataset~\cite{cellpose} is a comprehensive collection of 608 diverse cell images designed to evaluate segmentation algorithms across diverse conditions. This generalized dataset encompasses a wide range of cell types and imaging modalities, including fluorescent, brightfield, and non-microscopy images. Within this broader collection, a specialized subset of 100 images contains neurons with complex morphologies, presenting a particularly challenging test for segmentation methods.
    In our study, we utilized both subsets to evaluate segmentation algorithm performance across specific and diverse conditions.

    \textbf{Deepbacs}. The Deepbacs dataset~\cite{deepbacs, deepbacs_ecb, deepbacs_bsf} focuses on bacterial image segmentation. For our study, we concentrated on two subsets: Escherichia coli and Bacillus subtilis. Both species are rod-shaped bacteria, with different imaging methods. Escherichia coli was captured using bright field microscopy, while Bacillus subtilis was imaged using fluorescence microscopy. These subsets comprise time-lapse image sequences to showcase the bacterial growth process from a small initial population to a larger colony. We excluded subsets containing fewer than 10 images to ensure sufficient data for analysis.

    \textbf{TissueNet}. TissueNet~\cite{tissuenet} is a cell segmentation dataset containing over 1 million manually labeled cells from diverse tissue types across human, mouse, and macaque samples. It includes both normal and diseased tissues, imaged using various platforms such as CODEX, CyCIF and Vectra. We use nuclei annotations from 14 of the 19 TissueNet 1.1 subtypes, each containing more than 10 training images. These subtypes cover major tissue types including pancreas, immune, gastrointestinal, breast, and skin tissues, providing a broad representation of cellular morphologies and imaging conditions.

    \textbf{DSB2018}. The 2018 Data Science Bowl (DSB2018) dataset~\cite{dsb2018} comprises microscopy images from over 30 biological experiments, featuring 22 cell types at 15 different resolutions. It includes both fluorescence microscopy of cultured cells and brightfield microscopy of stained tissue samples. For our study, we utilized a curated subset of 497 fluorescence images, refined by the StarDist team~\cite{stardist} to remove labeling errors. This subset, consisting of 447 training and 50 testing images, serves as a benchmark for cell nuclei segmentation algorithms.

    \textbf{CellSeg-Blood}. This dataset is a subset derived from the CellSeg challenge dataset~\cite{NeurIPS-CellSeg, cellseg_dataset}, a multimodal cell segmentation benchmark. We extracted blood cell images to create the CellSeg-Blood subset, comprising 139 training and 14 test images with cell instance annotations. These images feature diverse blood cell types with varying colors and shapes, set against complex, textured backgrounds. By incorporating this unique subset into our study, we aim to challenge our model with novel cellular structures and intricate backgrounds typical of blood smears, while also expanding the range of cell morphologies examined in our overall analysis. This selection complements our existing datasets, allowing for a more comprehensive evaluation of the model's adaptability and performance across different biological contexts.

    \subsection{Preprocess}
    Our study encompassed datasets with diverse sizes and color profiles. To ensure consistency, we standardized all images by converting them to uint8 format and normalizing pixel values to the 0 to 255 range. We observed that StarDist performed poorly on high-resolution images. Therefore, for a fair comparison, we resized all images from the DeepBacs and CellSeg Blood datasets to $512 \times 512$ pixels.

    We implemented a patch extraction method akin to Cellpose, using $256 \times 256$  pixel patches with 50\% overlap. To fully utilize the GPU parallel processing capabilities, we ensured each batch contained at least 32 patches for both CellSeg1 and comparison methods. For datasets yielding fewer than 32 patches, we replicated existing patches $\left\lceil \frac{32}{n} \right\rceil$ times, maintaining our single-image training paradigm while optimizing computational efficiency.

    The data augmentation pipeline for CellSeg1 includes several classical random transformations applied to the image and its corresponding mask.
    These transformations include adjusting brightness and contrast with a brightness limit of 0.1 and a contrast limit of 0.1, flipping with a probability of 0.75, and applying random resized cropping.
    The random resized cropping is performed with a crop scale range of 0.3 to 1.0 and a crop aspect ratio range of 0.75 to 1.33.
    Additionally, the image is randomly shifted, scaled, and rotated with a scale limit range of -0.5 to 0.5 and a rotation probability of 0.8.

    \subsection{Modifications to SAM}
    Previous SAM-based cell segmentation approaches typically required high-end hardware such as 80GB NVIDIA A100 GPUs~\cite{cellsam,cellvit} for training. To enable efficient fine-tuning on more accessible GPUs~(\extfigref{fig:extended_figure_5}{b}), the input image resolution was reduced from $1024 \times 1024$ to $512 \times 512$ pixels to decrease memory usage, and bilinear interpolation was applied to the positional encodings to accommodate this change.

    To leverage the pre-trained features of the original SAM model, we froze all of its parameters and only applied LoRA to the query and value projections within the transformer layers of the prompt encoder and mask decoder. In all experiments, the LoRA rank was set to 4, and the LoRA dropout rate was set to 0.1. We opted to use only point prompts.

    The output from the first mask token was used as the final mask prediction, and we did not utilize SAM's multiple valid mask output functionality. The IoU scores outputted by SAM were used as cell probability, indicating the probability of each mask being a cell. During training, we maintained consistency with the original SAM model by using a mean squared error loss, with a target value of 1 for cells and 0 for non-cells.

    \subsection{Training strategy}
    During the training process, our model processes an image-point pair as input. The image is encoded to generate image embedding, while the point, sampled from either a cell mask (positive sample) or background (negative sample), is processed by a prompt encoder. These embeddings are then fed into a mask decoder, which produces a segmented mask and a cell probability for the input point.

    The loss function combines binary cross-entropy loss ($\mathcal{L}_{BCE}$)  for mask prediction and mean squared error loss ($\mathcal{L}_{MSE}$) for point classification. For negative samples, only $\mathcal{L}_{BCE}$ is calculated, while for positive samples, the final loss is the sum of $\mathcal{L}_{BCE}$ and $\mathcal{L}_{MSE}$. These losses are defined as:

    \begin{equation}
        \mathcal{L}_{BCE} = -\frac{1}{N} \sum_{i=1}^N \left[ y_i \log \left(\hat{y}_i\right) + \left(1-y_i\right) \log \left(1-\hat{y}_i\right) \right]
    \end{equation}

    \begin{equation}
        \mathcal{L}_{MSE} = \frac{1}{N} \sum_{i=1}^N \left(y_i-\hat{y}_i\right)^2
    \end{equation}

    where $N$ is the number of samples, $y_i$ is the ground truth label (1 for positive, 0 for negative), and $\hat{y}_i$ is the predicted cell probability.

    We employed the OneCycleLR~\cite{smith2019super} strategy, setting a maximum learning rate of 0.003 and a peak percentage of 0.3. The AdamW~\cite{loshchilov2017decoupled} optimizer was used with beta values of (0.9, 0.999), and a weight decay of 0.01. Training was conducted with a batch size of 4 and gradient accumulation factor of 8, resulting in an effective batch size of 32.~(\extfigref{fig:extended_figure_5}{})

    For CellSeg1 experiments using limited training data (1, 5, or 10 images), we set the number of epochs to 300. For complete datasets, we adjusted the training duration based on dataset size: 30 epochs for most datasets (CPG, CSB, CPS, BSF, and DSB), while the smaller ECB dataset (19 images) was trained for 100 epochs to ensure adequate convergence. Unless otherwise stated, all other CellSeg1's parameters remained identical across experiments.

    \subsection{Sampling strategy}
    Since our model requires an image-point pair as input during training, we need an effective strategy to sample representative points from cell masks. Random sampling of all pixels would be computationally inefficient and could introduce ambiguity near cell boundaries. Therefore, we developed a focused sampling approach that prioritizes informative regions while maintaining computational efficiency.

    For each image, we computed distance transforms for both foreground (cell) and background regions relative to cell boundaries. This allows us to identify areas furthest from edges, which are likely to be most representative of cell interiors and clear background. We then randomly sample points from the top 20\% of these distance-transformed regions, ensuring a focus on unambiguous areas while maintaining some variability.

    To maximize training efficiency, we processed multiple sample points from a single image simultaneously. This approach allows us to pass the image through the computationally intensive image encoder only once, while processing several dozen sample points through the lighter-weight prompt encoder and mask decoder. This strategy significantly reduces computational overhead, as the image encoder step is shared across all samples from the same image.

    The number of cells can vary significantly between images, so we limit the number of positive (cell) and negative (background) samples per image at 30 and 15, respectively. This allows us to process multiple images in a single batch while maintaining a consistent memory footprint.

    \subsection{Inference strategy}
    During inference, CellSeg1 leverages SAM's``everything" mode by uniformly sampling points on a $32 \times 32$ grid across the input image. For each point, the model generates a mask prediction and reinterprets SAM's original IoU output as a cell probability score, indicating whether the current point is located within a cell. Through an optimized NMS algorithm, these predictions are integrated to generate the final instance segmentation result.

    The inference process is computationally efficient since the image embedding only needs to be computed once through the encoder for all grid points. Given the lightweight nature of the prompt encoder and mask decoder, the prediction process takes approximately 2-5 seconds~(\extfigref{fig:extended_figure_3}{c}) per image on a single NVIDIA RTX 4090 GPU, with processing time varying depending on the number of cells present in the image.

    \subsection{Optimized mask NMS}
    The Non-Maximum Suppression (NMS) strategy plays a crucial role in generating accurate cell instance segmentation results. While the original SAM implementation uses box-based NMS for computational efficiency, we identified several limitations in cell segmentation scenarios that motivated our optimized approach.

    The conventional box-based NMS can lead to incorrect mask suppression when bounding boxes overlap despite their underlying cell masks being distinct~(\extfigref{fig:extended_figure_3}{a}). This is particularly problematic in densely packed cell environments where adjacent cells frequently have overlapping bounding boxes but separate boundaries.

    A potential solution is to perform mask-based NMS by directly calculating mask overlaps~(\extfigref{fig:extended_figure_3}{b}). However, this approach requires computing pairwise IoU scores between all mask pairs, resulting in quadratic computational complexity that becomes prohibitive for images with numerous cells.

    We use an optimized two-stage NMS strategy~(Algorithm \ref{alg:alg1}) that balances accuracy and computational efficiency~(\extfigref{fig:extended_figure_3}{c}). In the first stage, we apply the computationally efficient box-based NMS to identify potentially overlapping regions. In the second stage, we only compute mask IoU scores for those masks whose bounding boxes overlap, using an IoU threshold of 0.05, significantly reducing the number of required mask IoU calculations while maintaining high accuracy.

    Additionally, we improved the scoring mechanism for mask selection. While the original SAM uses a combination of box area reciprocal and predicted IoU as the mask score, our empirical studies showed that using the product of the predicted IoU and a stability score leads to more robust results in cell segmentation tasks. The stability score helps prioritize masks with well-defined and consistent boundaries, which is particularly important for accurate cell instance segmentation.

    \begin{algorithm}[H]
        \SetKwInOut{Input}{Input}
        \SetKwInOut{Output}{Output}
        \Input{$\mathcal{B} = \{b_1, \ldots, b_N\}$: set of bounding boxes\\
            $\mathcal{M} = \{m_1, \ldots, m_N\}$: set of masks\\
            $\mathcal{S} = \{s_1, \ldots, s_N\}$: set of scores\\
            $\tau$: IoU threshold}
        \Output{$\mathcal{D}$: set of kept detections}

        $\mathcal{D} \leftarrow \emptyset$

        \tcp{$\mathbf{O}$: binary matrix where $O_{ij} = 1$ if boxes i and j overlap, 0 otherwise}

        $\mathbf{O} \leftarrow \text{ComputeBoxOverlapMatrix}(\mathcal{B})$

        $I \leftarrow \text{argsort}(\mathcal{S}, \text{descending=True})$

        % \tcp{$I$: list of indices sorted by descending scores}

        \While{$I \neq \emptyset$}{
            $k \leftarrow \text{Pop}(I)$

            $\mathcal{D} \leftarrow \mathcal{D} \cup \{k\}$

            $J \leftarrow \{i \in I : \mathbf{O}_{ki} = 1\}$

            \ForEach{$i \in J$}{
                $iou \leftarrow \text{ComputeMaskIoU}(m_k, m_i)$

                \If{$iou > \tau$}{
                    $I \leftarrow I \setminus \{i\}$
                }
            }
        }
        \Return{$\mathcal{D}$}
        \caption{Optimized Mask Non-Maximum Suppression (NMS)}
        \label{alg:alg1}
    \end{algorithm}

    \subsection{Benchmarking}
    For all experiments, Cellpose and StarDist were trained following the optimal hyperparameters reported in the Cellpose paper. Cellpose was trained for 500 epochs with an initial learning rate of 0.2, while StarDist was trained for 1000 epochs with an initial learning rate of 0.0007. To ensure sufficient training iterations for datasets with fewer than 32 patches, we replicated the patches to reach at least 32, allowing the epochs to be effectively utilized even for a single image.

    To ensure consistency, CellSeg1 and Cellpose employed identical patch extraction methods, using a 256x256 patch size with 50\% overlap. StarDist also used a 256x256 patch size but relied on its default random sampling-based patch extraction method.

    Each method incorporated its respective data augmentation techniques. Cellpose applied its default augmentations, including random rotations, scalings, and translations. We turn off cellpose's size resizing due to the discrepancies between single image and dataset-wide average cell sizes. StarDist's augmentations encompassed random flips, rotations, pixel intensity adjustments, and the addition of random Gaussian noise.

    We processed image colors differently across methods. SAM, StarDist and CellSeg1 directly process RGB images, while Cellpose and CellSAM require channel-specific inputs. For these channel-specific models, we adopt a flexible strategy~(Supplementary Table 2) - using single-channel input for grayscale datasets (like DSB, ECB, BSF), while for multi-channel datasets (like CPS, TSN), selectively utilizing main and secondary channels based on nuclei and cytoplasmic distribution to maximize cellular morphological information. RGB images (like CSB) are converted to grayscale to maintain processing consistency.

    All other parameters not explicitly mentioned for Cellpose and StarDist were kept at their default settings.
    \subsection{Metric}
    For evaluating segmentation performance, we adopted the same metric as Cellpose, using mean Average Precision (mAP) at 0.5 IoU thresholds.

    The evaluation process begins by matching each predicted mask with its closest ground-truth mask based on IoU scores. True positives (TP) are defined as predicted masks that successfully match a ground-truth mask above the IoU threshold, false positives (FP) are predicted masks without valid matches, and false negatives (FN) are ground-truth masks that remain unmatched. The Average Precision (AP) for each image is then calculated as:

    \begin{equation}
        AP = \frac{TP}{TP+FP+FN}
    \end{equation}

    The final mAP score is computed by averaging the AP values across all images in the test set.

    \subsection{Implementation}
    Our codebase is implemented in Python, built on PyTorch (\url{https://pytorch.org/}) with the Albumentations~\cite{albumentations} library for data augmentation. We employed the ViT-H variant of the SAM model across all experiments, which demonstrated superior performance compared to other model sizes~(\extfigref{fig:extended_figure_4}{}). All hyperparameters were kept identical across experiments, with the sole exception being the number of training epochs when training on complete datasets.
    All experiments were conducted on RTX 4090 GPUs using Ray (\url{https://www.ray.io/}) for parallel execution. While we ran multiple experiments simultaneously for efficiency, each experiment requires only a single GPU.

    We developed a graphical user interface using Streamlit (\url{https://streamlit.io/}) and Plotly (\url{https://plotly.com/}), allowing users to perform training, testing, and visualization through an intuitive interface. All figures presenting segmentation results in this paper were generated using Plotly.

    \subsection{Reproducibility}
    To achieve exact reproducibility of our study, we eliminated all sources of randomness by fixing the random seed to 0 and disabling any non-deterministic operations during both training and testing phases. All CellSeg1's segmentation results and their visualization figures in this study can be precisely reproduced using our publicly available codebase with identical environment settings and datasets. To facilitate exact replication, we have provided detailed documentation covering environment setup and hyperparameter specifications in our GitHub repository, allowing researchers to obtain identical results.

    \subsection{Data availability}
    The Data Science Bowl 2018 (DSB) dataset is available from the StarDist repository at \url{https://github.com/stardist/stardist/releases/download/0.1.0/dsb2018.zip}. The Cellpose datasets (CPS, CPG) were sourced from \url{https://www.cellpose.org/dataset}. DeepBacs bacterial datasets for E. coli brightfield (ECB) and B. subtilis fluorescence (BSF) are available from Zenodo at \url{https://zenodo.org/records/5550935} and \url{https://zenodo.org/records/5550968}, respectively. The CellSeg-Blood (CSB) dataset was obtained from the CellSeg challenge dataset available at \url{https://zenodo.org/records/10719375}. TissueNet (TSN) nuclei dataset was sourced from the DeepCell project at \url{https://datasets.deepcell.org/data}. All processed code used in this study are available in our GitHub repository at \url{https://github.com/Nuisal/cellseg1}.

    \subsection{Code availability}
    The code is available at \url{https://github.com/Nuisal/cellseg1}. The repository contains the complete implementation of CellSeg1, including a graphical user interface for training, testing and visualization, data preprocessing pipelines, and documentation. The codebase includes scripts for reproducing CellSeg1's results, evaluating CellSAM and zero-shot SAM performances, and generating the figures presented in this paper (except Figure 2).

    \subsection{Acknowledgements}
    We thank all researchers for sharing their cell image datasets openly. Special thanks to Meta AI for open-sourcing the SAM, making it accessible to researchers worldwide.
    \subsection{Competing interests}
    The authors declare no competing interests.
    \subsection{Author contributions}
    Y.X, B.D and P.Z conceived the study. P.Z performed experiments, analyzed data, wrote the code and manuscript. Y.X and B.D analyzed data and edited the manuscript.

    \section{Extended data figure}
    \renewcommand{\figurename}{Extended Data Fig.}
    \setcounter{figure}{0}

    \begin{figure}[H]
        \centering
        \begin{tikzpicture}
  \node[anchor=north west, xshift=0, yshift= 00.00cm] (row_1) at (0,0) {\includestandalone{figures/extended_figure_1/extended_figure_1_1}};
  \node[anchor=north west, xshift=0, yshift=-03.15cm] (row_2) at (0,0) {\includestandalone{figures/extended_figure_1/extended_figure_1_2}};
  \node[anchor=north west, xshift=0, yshift=-06.30cm] (row_3) at (0,0) {\includestandalone{figures/extended_figure_1/extended_figure_1_3}};
  \node[anchor=north west, xshift=0, yshift=-09.45cm] (row_4) at (0,0) {\includestandalone{figures/extended_figure_1/extended_figure_1_4}};
  \node[anchor=north west, xshift=0, yshift=-12.60cm] (row_5) at (0,0) {\includestandalone{figures/extended_figure_1/extended_figure_1_5}};

  \node[anchor=north west, rotate=00, xshift=000.00cm, yshift=0.0000cm] at (0,0) {\textbf{a}};
  \node[anchor=north west, rotate=90, xshift=-02.20cm, yshift=-0.00cm] at (0,0) {CSB};
  \node[anchor=north west, rotate=00, xshift=000.00cm, yshift=-03.15cm] at (0,0) {\textbf{b}};
  \node[anchor=north west, rotate=90, xshift=-5.30cm, yshift=-0.00cm] at (0,0) {ECB};
  \node[anchor=north west, rotate=00, xshift=000.00cm, yshift=-06.30cm] at (0,0) {\textbf{c}};
  \node[anchor=north west, rotate=90, xshift=-8.30cm, yshift=-0.00cm] at (0,0) {BSF};
  \node[anchor=north west, rotate=00, xshift=000.00cm, yshift=-09.45cm] at (0,0) {\textbf{d}};
  \node[anchor=north west, rotate=90, xshift=-11.50cm, yshift=-0.00cm] at (0,0) {DSB};
  \node[anchor=north west, rotate=00, xshift=000.00cm, yshift=-12.60cm] at (0,0) {\textbf{e}};
  \node[anchor=north west, rotate=90, xshift=-14.75cm, yshift=-0.00cm] at (0,0) {CPS};
\end{tikzpicture}
        \caption{Impact of annotation quality and cell count on CellSeg1 performance across diverse datasets. Each row represents a different dataset, with images arranged from left to right by increasing cell count. The top-left of each image shows the $mAP_{0.5}$ achieved on the test set when using that single image for training, while the top-right indicates the number of annotated cells. Yellow outlines in the bottom-right quadrant represent cell annotations.}
        \label{fig:extended_figure_1}
    \end{figure}
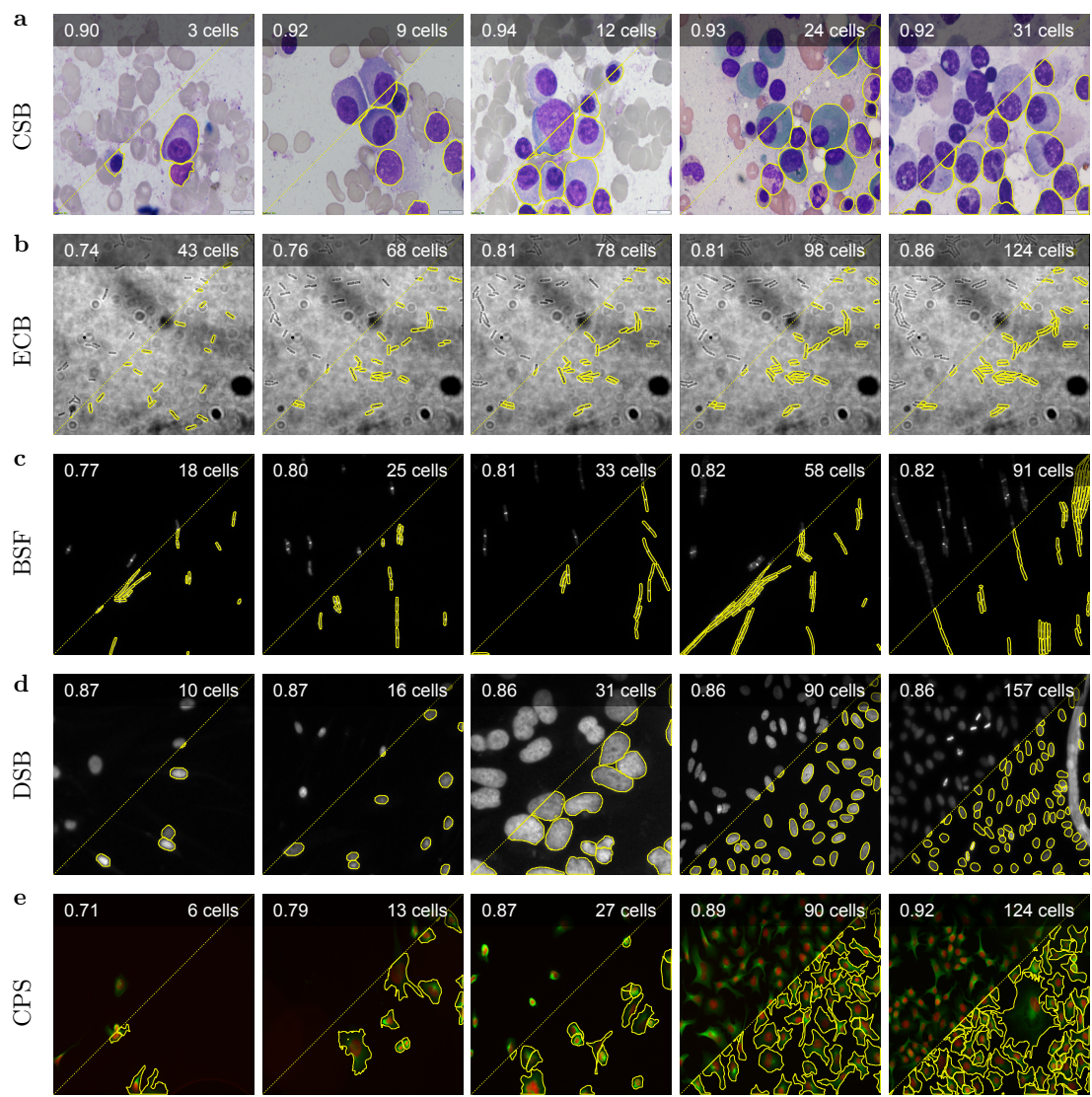

    \begin{figure}[H]
        \centering
        \begin{tikzpicture}
  \node[anchor=north west, xshift=0, yshift= 00.00cm] (row_1) at (0,0) {\includestandalone{figures/extended_figure_2/extended_figure_2_1}};
  \node[anchor=north west, xshift=0, yshift=-3.75cm] (row_2) at (0,0) {\includestandalone{figures/extended_figure_2/extended_figure_2_2}};
  \node[anchor=north west, xshift=0, yshift=-7.5cm] (row_3) at (0,0) {\includestandalone{figures/extended_figure_2/extended_figure_2_3}};
  \node[anchor=north west, xshift=0, yshift=-11.25cm] (row_4) at (0,0) {\includestandalone{figures/extended_figure_2/extended_figure_2_4}};
  \node[anchor=north west, xshift=0, yshift=-15cm] (row_5) at (0,0) {\includestandalone{figures/extended_figure_2/extended_figure_2_5}};

  \draw[dashed] (4.44cm,-0.15cm) -- (4.44cm,-18.75cm);
  \node[anchor=north west, rotate=00, xshift=1.75cm, yshift=0.4cm] at (0,0) {Train};
  \node[anchor=north west, rotate=00, xshift=5.9cm, yshift=0.4cm] at (0,0) {Test};
  \node[anchor=north west, rotate=00, xshift=9.3cm, yshift=0.4cm] at (0,0) {\textcolor[HTML]{FF6188}{Red} Box};
  \node[anchor=north west, rotate=00, xshift=13.0cm, yshift=0.4cm] at (0,0) {\textcolor[HTML]{A9DC76}{Green} Box};

  \node[anchor=north west, rotate=00, xshift=0.40cm, yshift=0.35cm] at (0,0) {\textbf{a}};
  \node[anchor=north west, rotate=00, xshift=4.60cm, yshift=0.4cm] at (0,0) {\textbf{b}};
  \node[anchor=north west, rotate=00, xshift=8.40cm, yshift=0.35cm] at (0,0) {\textbf{c}};
  \node[anchor=north west, rotate=00, xshift=12.1cm, yshift=0.4cm] at (0,0) {\textbf{d}};

  \node[anchor=north west, rotate=90, xshift=-02.5cm, yshift=000.00cm] at (0,0) {CSB};
  \node[anchor=north west, rotate=90, xshift=-6.2cm, yshift=000.00cm] at (0,0) {ECB};
  \node[anchor=north west, rotate=90, xshift=-9.9cm, yshift=000.00cm] at (0,0) {BSF};
  \node[anchor=north west, rotate=90, xshift=-13.7cm, yshift=000.00cm] at (0,0) {DSB};
  \node[anchor=north west, rotate=90, xshift=-17.5cm, yshift=000.00cm] at (0,0) {CPS};
\end{tikzpicture}
        \caption{CellSeg1's performance using single-image training across diverse cell types.
            (a) High-quality training images, which are the same exemplary images shown in Figure 3g, with their corresponding $mAP_{0.5}$ scores on the test set and cell counts. Yellow outlines indicate ground truth annotations.
            (b) Test images with predicted cell masks (cyan outlines) and their $AP_{0.5}$ scores. (c,d) Zoomed-in views of regions marked by red and blue boxes in (b), respectively, showing detailed segmentation results.}
        \label{fig:extended_figure_2}
    \end{figure}
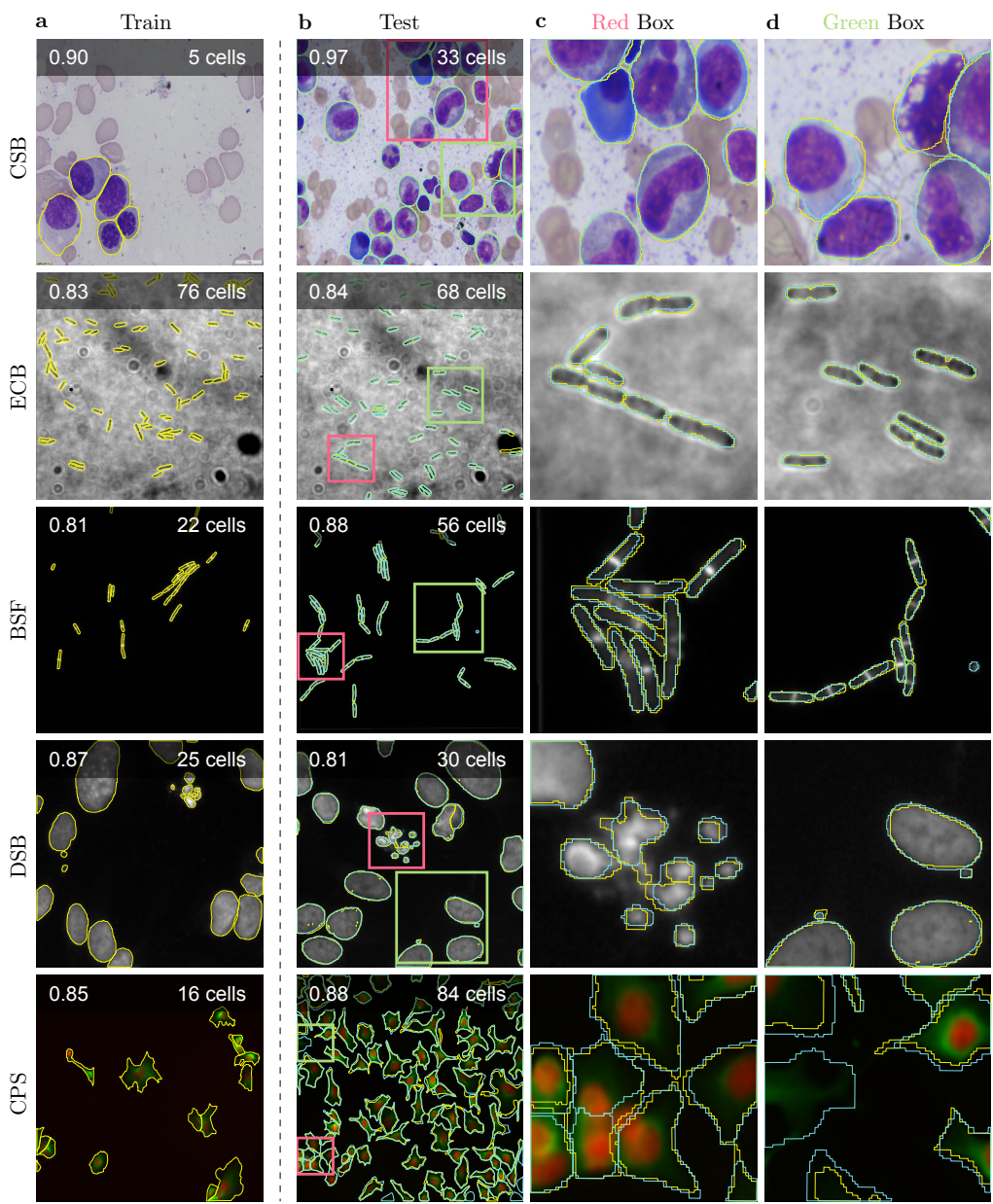

    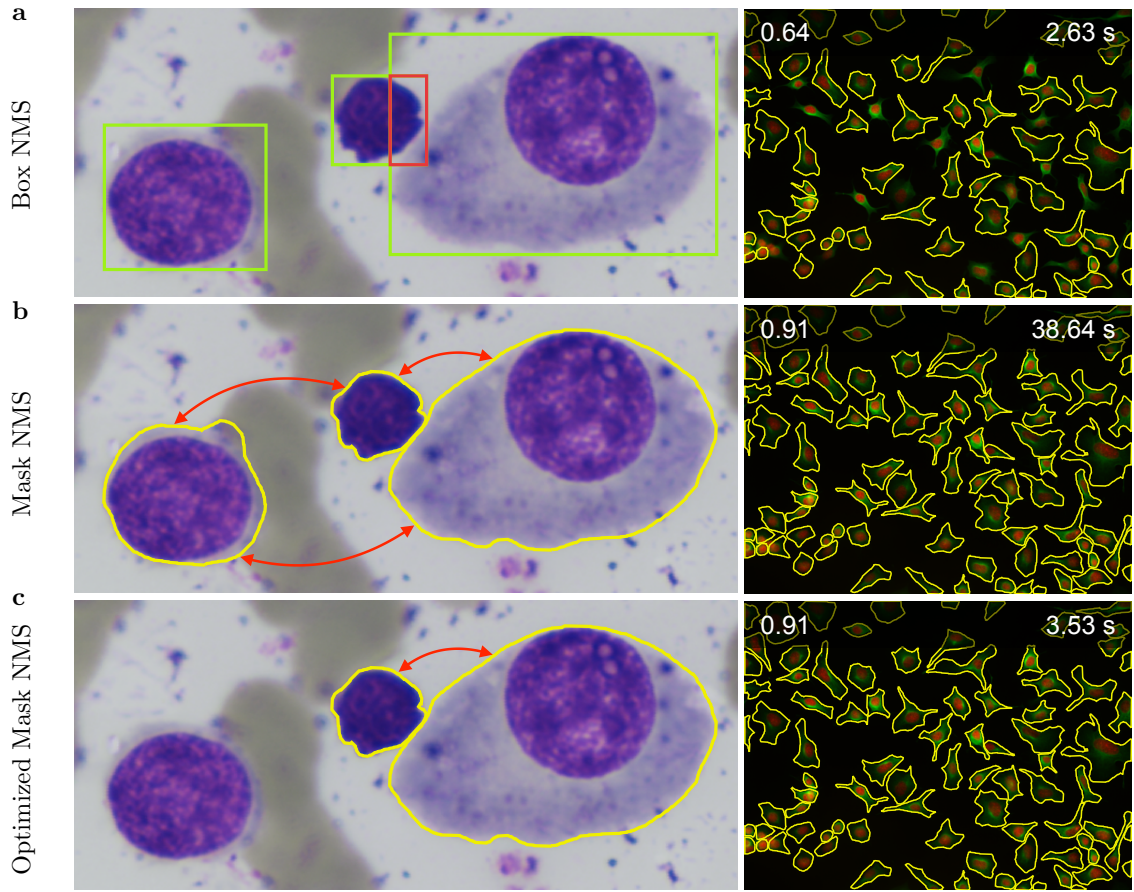
\begin{figure}[H]
        \centering
        \begin{tikzpicture}[font=\scriptsize]
  \node[anchor=north west, rotate=00, xshift=00.00cm, yshift=0.0000cm] (box_nms) at (0,0) {\includestandalone{figures/extended_figure_3/extended_figure_3_1}};
  \node[anchor=north west, rotate=00, xshift=00.00cm, yshift=-04.05cm] (mask_nms) at (0,0) {\includestandalone{figures/extended_figure_3/extended_figure_3_2}};
  \node[anchor=north west, rotate=00, xshift=00.00cm, yshift=-08.10cm] (opt_nms) at (0,0) {\includestandalone{figures/extended_figure_3/extended_figure_3_3}};
  \node[anchor=north west, xshift=0cm, yshift=0cm] at (0,0) {\textbf{\normalsize{a}}};
  \node[anchor=north west, xshift=0cm, yshift=-4cm] at (0,0) {\textbf{\normalsize{b}}};
  \node[anchor=north west, xshift=0cm, yshift=-8cm] at (0,0) {\textbf{\normalsize{c}}};
  \node[anchor=north west, rotate=90, xshift=-3.0cm, yshift=-0.0cm] at (0,0) {\normalsize{Box NMS}};
  \node[anchor=north west, rotate=90, xshift=-7.2cm, yshift=-0.0cm] at (0,0) {\normalsize{Mask NMS}};
  \node[anchor=north west, rotate=90, xshift=-12.1cm, yshift=-0.0cm] at (0,0) {\normalsize{Optimized Mask NMS}};
\end{tikzpicture}
        \caption{Comparison of three NMS strategies for cell segmentation. (a) Box-based NMS used in SAM relies on bounding box overlap for mask suppression. While computationally efficient, this approach may incorrectly suppress masks when bounding boxes overlap despite cells being separate, leading to over-suppression and reduced accuracy. The right panel shows the segmentation results, with $AP_{0.5}$ score in the top-left corner and processing time (averaged over 10 runs) in the top-right corner. (b) Pure mask-based NMS uses direct mask overlap calculations (indicated by red arrows) for more accurate results. While this achieves better segmentation accuracy, it requires computationally expensive pairwise IoU calculations between all masks, resulting in significantly longer processing time. (c) Our optimized two-stage approach combines the advantages of both methods: it first uses fast box-based NMS to identify potential overlaps, then only computes mask IoU for these overlapping regions. This strategy achieves the same high accuracy as pure mask-based NMS while maintaining computational efficiency close to box-based NMS.}
        \label{fig:extended_figure_3}
    \end{figure}

    \begin{figure}[H]
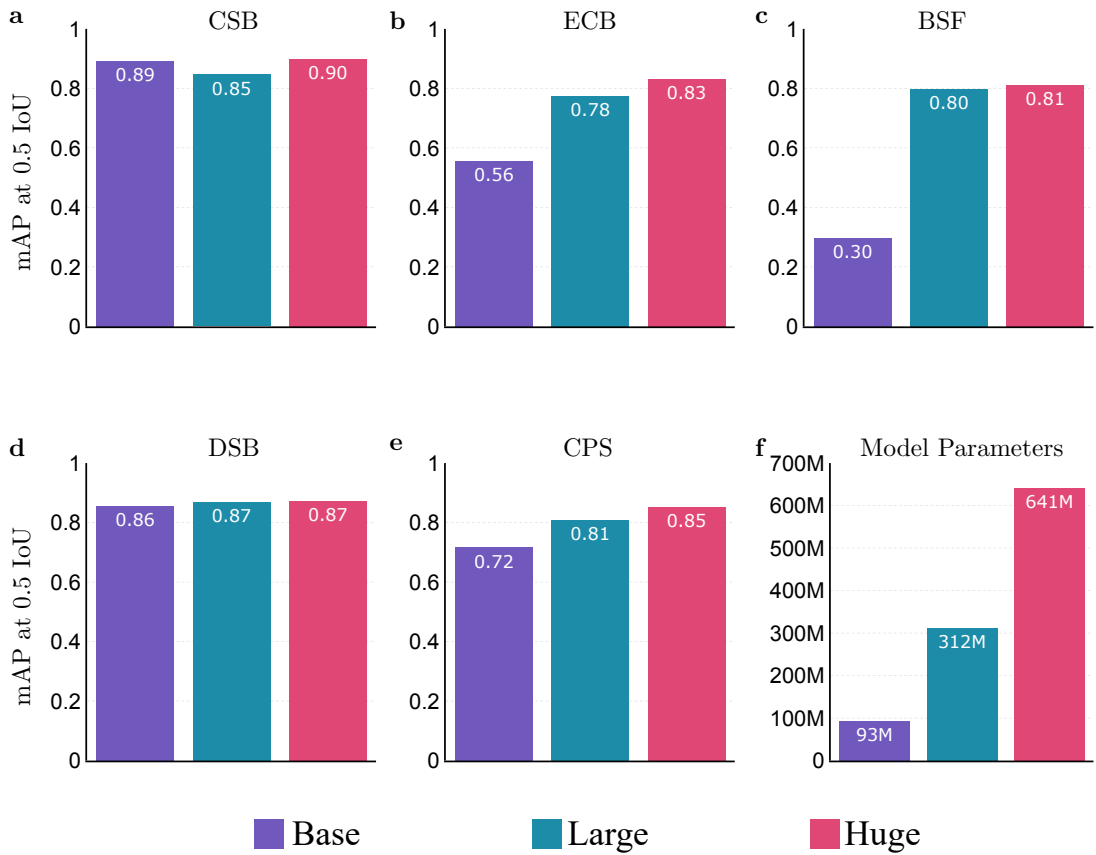

        \centering
        \includestandalone[width=1.0\textwidth]{figures/extended_figure_4/extended_figure_4}
        \caption{Performance comparison of SAM models with different model sizes. (a-e) $mAP_{0.5}$ IoU scores for each model variant when trained on the same single image shown in \figref{fig:figure_3}{g}. While larger models generally achieve better segmentation performance, the base model shows notably poor results on challenging datasets like ECB, CPS and BSF. (f) Number of parameters for each model size.}
        \label{fig:extended_figure_4}
    \end{figure}

    \begin{figure}[H]
        \centering
        \includestandalone[width=1.0\textwidth]{figures/extended_figure_5/extended_figure_5}
        \caption{Impact of batch size on model training efficiency and performance. (a) Training time (minutes), (b) peak GPU memory usage (GB), and (c) segmentation performance ($mAP_{0.5}$) across different datasets under various batch sizes (1, 2, and 4) on an NVIDIA RTX 4090. Gradient accumulation was used to maintain an effective batch size of 32 across all experiments (e.g., batch size 2 uses 16 accumulation steps). All experiments were conducted using the high-quality training images shown in \figref{fig:figure_3}{g}.}
        \label{fig:extended_figure_5}
    \end{figure}

    \begin{figure}[H]
        \centering
        \includegraphics[width=1.0\textwidth]{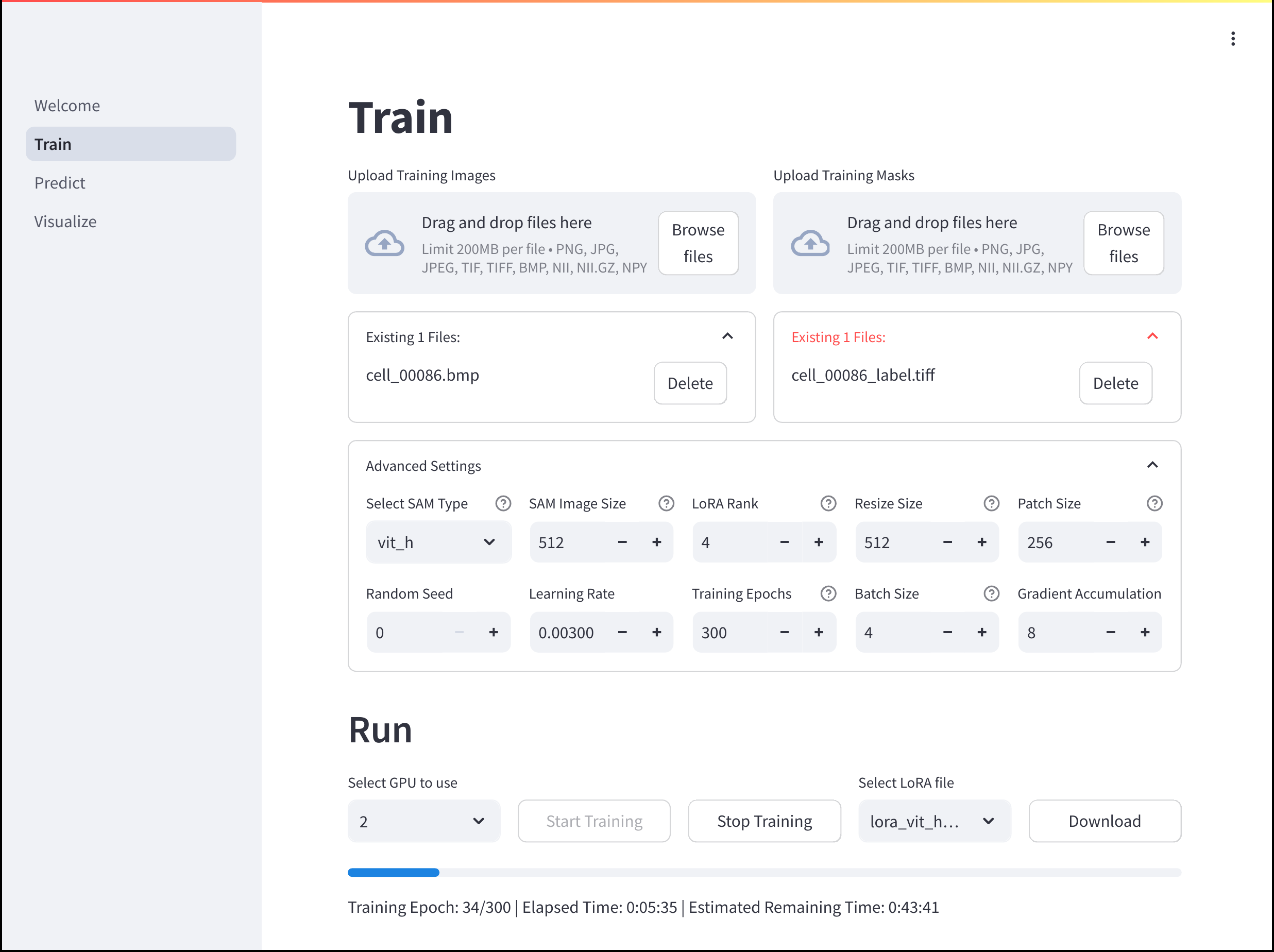}
        \caption{A web-based graphical interface for CellSeg1, implemented with Streamlit, enabling flexible deployment where the training server and client can be on different machines. The interface provides functionalities through its sidebar menu, including model training, inference, and result visualization. The training interface shown here features intuitive controls for uploading training images and annotations, GPU selection, training progress monitoring.}
        \label{fig:extended_figure_6}
    \end{figure}

    \section{Supplementary Table}
    \renewcommand{\tablename}{Supplementary Table}
    \begin{table}[H]
        \centering
        \begin{tabular}{|c|c|c|c|c|}
            \hline
            \multirow{2}{*}{Dataset name}      & \multirow{2}{*}{Subset}           & \multirow{2}{*}{Abbr.} & \multicolumn{2}{c|}{Image numbers}        \\
            \cline{4-5}
                                               &                                   &                        & Train                              & Test \\
            \hline
            \multirow{2}{*}{Cellpose}          & Specialized                       & CPS                    & 89                                 & 11   \\
            \cline{2-5}
                                               & Generalized                       & CPG                    & 540                                & 68   \\
            \hline
            DSB2018                            & StarDist                          & DSB                    & 447                                & 50   \\
            \hline
            CellSeg                            & Blood cell                        & CSB                    & 139                                & 14   \\
            \hline
            \multirow{2}{*}{Deepbacs}          & Escherichia coli, bright field    & ECB                    & 19                                 & 15   \\
            \cline{2-5}
                                               & Bacillus subtilis, fluorescence   & BSF                    & 80                                 & 10   \\
            \hline
            \multirow{14}{*}{Tissuenet nuclei} & Breast, 2019-12-11                & TSN-1                  & 32                                 & 16   \\
            \cline{2-5}
                                               & Breast, 2020-01-16                & TSN-2                  & 73                                 & 40   \\
            \cline{2-5}
                                               & Breast, 2020-05-26                & TSN-3                  & 268                                & 136  \\
            \cline{2-5}
                                               & Epidermis, 2020-02-26             & TSN-4                  & 67                                 & 36   \\
            \cline{2-5}
                                               & Epidermis, 2020-06-23             & TSN-5                  & 19                                 & 12   \\
            \cline{2-5}
                                               & Colon, 2019-12-19                 & TSN-6                  & 184                                & 92   \\
            \cline{2-5}
                                               & Esophagus, 2020-02-19             & TSN-7                  & 333                                & 168  \\
            \cline{2-5}
                                               & Colon, 2020-06-27                 & TSN-8                  & 268                                & 136  \\
            \cline{2-5}
                                               & lymph node metastasis, 2020-02-10 & TSN-9                  & 295                                & 148  \\
            \cline{2-5}
                                               & Lymph node, 2020-01-14            & TSN-10                 & 60                                 & 32   \\
            \cline{2-5}
                                               & Lymph node, 2020-05-20            & TSN-11                 & 19                                 & 12   \\
            \cline{2-5}
                                               & Pancreas, 2020-05-12              & TSN-12                 & 19                                 & 12   \\
            \cline{2-5}
                                               & Pancreas, 2020-06-24              & TSN-13                 & 587                                & 296  \\
            \cline{2-5}
                                               & Tonsil, 2020-02-11                & TSN-14                 & 335                                & 168  \\
            \hline
        \end{tabular}
        \caption{Overview of datasets used in this study. The table presents 5 main datasets with their respective subsets, abbreviations, and data split statistics.}
        \label{table:table1}
    \end{table}

    \begin{table}[H]
        \centering
        \begin{tabular}{|c|c|c|c|c|c|c|c|}
            \hline
            \multicolumn{3}{|c|}{Original Color Channels} & {CellSeg1}                                 & \multicolumn{2}{c|}{Cellpose}   & \multicolumn{2}{c|}{CellSAM}                                                     \\
            \cline{1-3} \cline{5-8}
            Dataset                                       & Nuclei                                     & Cytoplasm                       & and StarDist                 & Main  & Secondary             & Main  & Secondary \\
            \hline
            DSB                                           & \multicolumn{2}{c|}{\multirow{3}{*}{Gray}} & \multirow{3}{*}{3 Channel gray} & \multirow{3}{*}{Gray}        &       & \multirow{3}{*}{Gray} &                   \\
            \cline{1-1}
            ECB                                           & \multicolumn{2}{c|}{}                      &                                 &                              &       &                       &                   \\
            \cline{1-1}
            BSF                                           & \multicolumn{2}{c|}{}                      &                                 &                              &       &                       &                   \\
            \hline
            CPS                                           & Red                                        & Green                           & RGB                          & Green & Red                   & Green & Red       \\
            \hline
            TSN                                           & Green                                      & Blue                            & RGB                          & Green & Blue                  & Green &           \\
            \hline
            CSB                                           & \multicolumn{2}{c|}{RGB}                   & RGB                             & Gray*                        &       & Gray*                 &                   \\
            \hline
            \multirow{2}{*}{CPG}                          & \multicolumn{2}{c|}{Gray}                  & 3 Channel gray                  & Gray                         &       & Gray                  &                   \\
            \cline{2-8}
                                                          & Red                                        & Green                           & RGB                          & Gray* &                       & Green & Red       \\
            \hline
        \end{tabular}
        \caption{Dataset color configurations. DSB, ECB, and BSF datasets contain grayscale images used as the main channel for both models. For the CSB dataset, which consists of RGB images where nuclei and cytoplasmic content are not separated into distinct channels, both models convert the images to grayscale (indicated as gray* in the table) by averaging across channels following Cellpose's implementation. For Cellpose training, we utilized the Blue channel as a secondary input on the TSN dataset to maximize information usage. The CPG dataset contains two types of images: grayscale images and two-channel fluorescence images with separate nuclear (Red) and cytoplasmic (Green) channels. For consistency in training, Cellpose converts all CPG images to grayscale, regardless of their original format. Through empirical testing, we found that CellSAM performs optimally on TSN without using the secondary channel, while benefiting from the secondary channel on CPG (for two-channel images) and CPS. The remaining dataset configurations are shown directly in the table, where nuclei and Cytoplasm columns indicate the original image channels, while Main and Secondary columns show how these channels are used by each model.}
        \label{table:table2}
    \end{table}

\bibliographystyle{naturemag}
\bibliography{ref.bib}

\end{document}